\documentclass[letterpaper, 10 pt, conference]{ieeeconf}  % Comment this line out if you need a4paper
\IEEEoverridecommandlockouts                              % This command is only needed if 
\usepackage{graphics} % for pdf, bitmapped graphics files
\usepackage{epsfig} % for postscript graphics files
\usepackage{mathptmx} % assumes new font selection scheme installed
\usepackage{amsmath} % assumes amsmath package installed
\usepackage{amssymb}  % assumes amsmath package installed

\usepackage[]{algorithm2e}
\usepackage[dvipsnames]{xcolor}
\usepackage{subcaption}
\usepackage[normalem]{ulem}

\title{\LARGE \bf
Virtual model control for compliant reaching under uncertainties
}

\author{Yi Zhang$^{1}$, Daniel Larby$^{1}$, Fumiya Iida$^{1}$, and Fulvio Forni$^{1}$% <-this % stops a space
\thanks{*Y. Zhang is supported by Sciences Research Council and AgriFoRwArdS CDT EP/S023917/1.
D. Larby is partially supported by the EPSRC grant EP/T517847/1 and by CMR Surgical.}
\thanks{$^{1}$Department of Engineering, 
        University of Cambridge, UK.}
}

\begin{document}

\maketitle
\thispagestyle{empty}
\pagestyle{empty}

%%%%%%%%%%%%%%%%%%%%%%%%%%%%%%%%%%%%%%%%%%%%%%%%%%%%%%%%%%%%%%%%%%%%%%%%%%%%%%%%
\begin{abstract}

Virtual Model Control (VMC) is an approach to design a controller for force-controlled robots in complex uncertain environments. While this method was primarily investigated for legged robot locomotion in the past, it can be more generally applicable to other types of robotic systems. This paper investigates the VMC framework for reaching tasks in a force-controlled robotic arm. We propose six different approaches to designing virtual models in order to achieve reaching tasks in environments with obstacles and uncertainties. A force-controlled 8 degree-of-freedom humanoid robot was used to validate the proposed approach in the real world. We conducted three experiments to test the performance of VMC controllers in terms of predictability, sensitivity to external force, and adaptability against known and unknown obstacles. Experimental analyses show that, even though the proposed approach needs to sacrifice accuracy and trajectory optimality, it enables us to design complex reaching motions under uncertainties, in an intuitive and extendable manner.
\end{abstract}

%%%%%%%%%%%%%%%%%%%%%%%%%%%%%%%%%%%%%%%%%%%%%%%%%%%%%%%%%%%%%%%%%%%%%%%%%%%%%%%%
\section{Introduction}

Safe planning and control of robotic arms for reaching tasks in uncertain cluttered environment are still unsolved problems. 
The fundamental issue lies in the inability of traditional motion control and planning methods to negotiate in real time with real-world scenarios, which are inherently unstructured and where object shapes and obstacle locations are only partially known. 
In fact, the remarkable speed and accuracy of our best industrial robots require a well-known and restricted factory floor environment.

The conventional solutions typically employ an interplay of complex sensing, state estimation, planning, and feedback control, but most of them require complex and computationally expensive sensing and planning processes, which lack interact-ability to accommodate uncertainties.
Furthermore, the best planning methods such as Probabilistic Road Map (PRM) \cite{LaValle1998RapidlyexploringRT}, Rapidly Exploring Random Tree (RRT) \cite{508439}, Potential field (PF) \cite{apf} are then often realised through inverse kinematics and joint-space high-gain PID control. This makes the robot stiff to guarantee accurate tracking, at the cost of low sensitivity to external perturbations, such as unknown obstacles.
As a result, conventional solutions show limitations in cluttered environments, characterized by inaccurate information. 

This paper advocates for methods that integrate motion planning and motion control to handle uncertainties in a safe way. Ideally, we want to take advantage of the available information on the environment and also offer a certain degree of compliant motion for real-time adaptation to unexpected scenarios. 
We propose the use of \emph{virtual model control} \cite{Pratt2001VirtualMC} for the safe, intuitive, and tractable approach to plan and control reaching tasks for articulated robotic arms.
In virtual model control, the robot behavior emerges from the interaction between the mechanics of the robot interconnected with designed `virtual' mechanism, given by a collection of springs and dampers organized in a suitable geometry. 
Eventually, this `embodied’ controller is simulated in a computer and realized through current/torque commands to the actuators. 
The virtual mechanism pulls the robot behavior into specific patterns, leaving a certain degree of sensitivity to the interaction with the environment. 
The robot steers in the direction defined by the total energy of the system, shaped by virtual mechanisms and environment. Therefore, trajectory planning and motion control are integrated. 

\begin{figure*}[ht]
	\centering
	\begin{subfigure}[h]{0.33\textwidth}
		\centering
		\includegraphics[width=.6\textwidth]{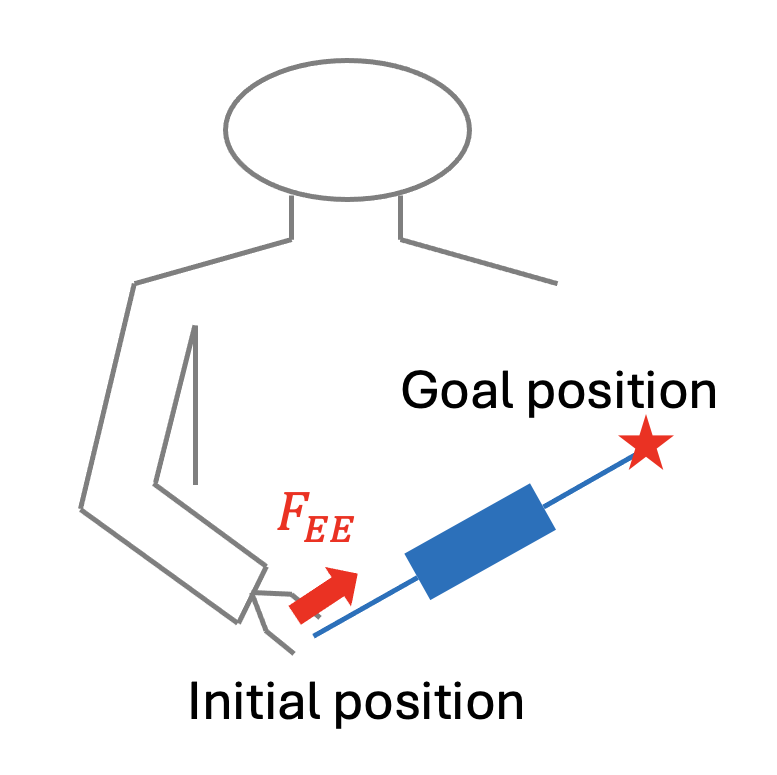}
		\caption{Simple reaching.}
		\label{goal_reaching}
	\end{subfigure}\hfill
	\begin{subfigure}[h]{0.33\textwidth}
		\centering
        \vspace{3mm}
		\includegraphics[width=.6\textwidth]{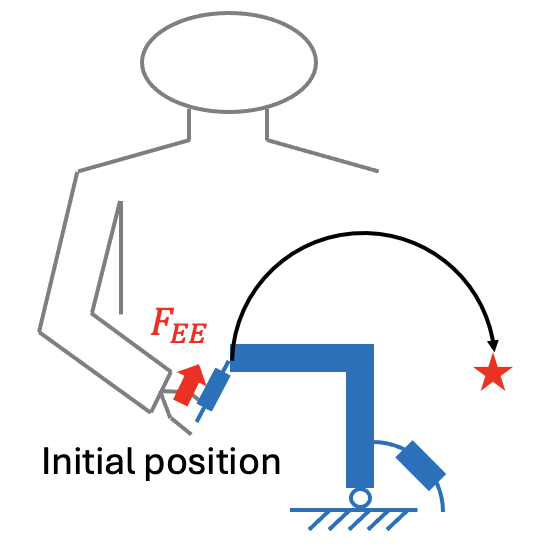}
		\caption{Geometrical reaching.}
		\label{goal_reaching_complicated}
	\end{subfigure}\hfill
	\begin{subfigure}[h]{0.33\textwidth}
		\centering
		\includegraphics[width=.6\textwidth]{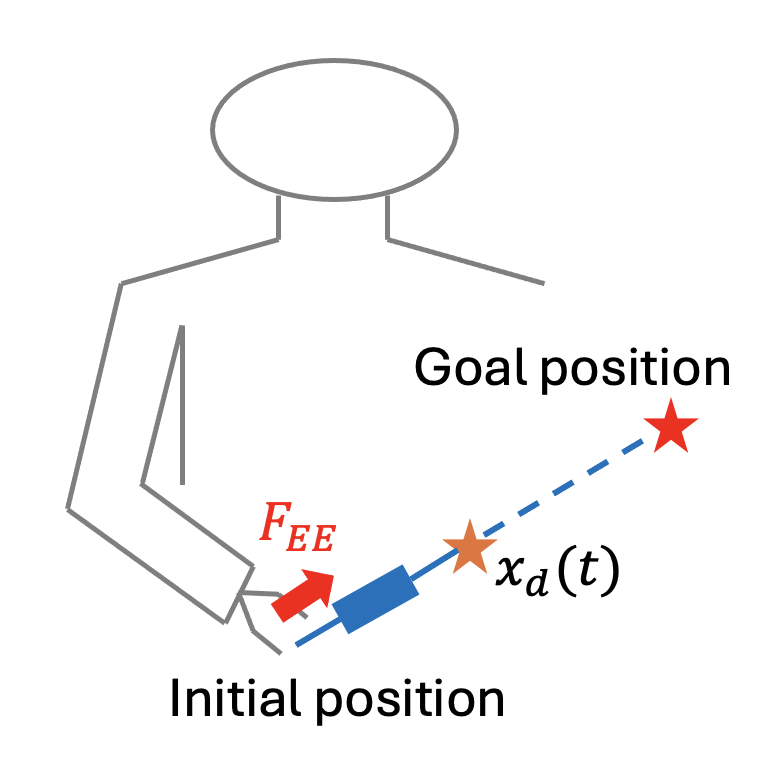}
		\caption{Moving-reference reaching.}
		\label{trajectory}
	\end{subfigure}\hfill
	\begin{subfigure}[h]{0.33\textwidth}
		\centering
		\includegraphics[width=.6\textwidth]{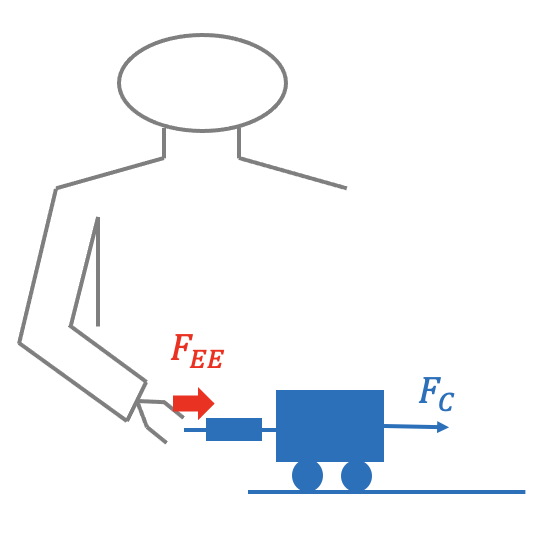}
		\caption{Force-driven reaching.}
		\label{trajectory_complicated}
	\end{subfigure}\hfill
	\begin{subfigure}[h]{0.33\textwidth}
		\centering
		\includegraphics[width=.6\textwidth]{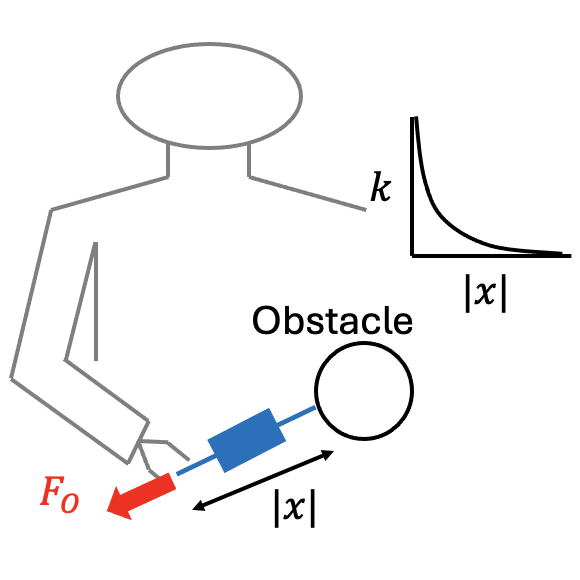}
		\caption{Obstacle avoidance.}
		\label{obstacle_avoidance1}
	\end{subfigure}\hfill
	\begin{subfigure}[h]{0.33\textwidth}
		\centering
		\includegraphics[width=.6\textwidth]{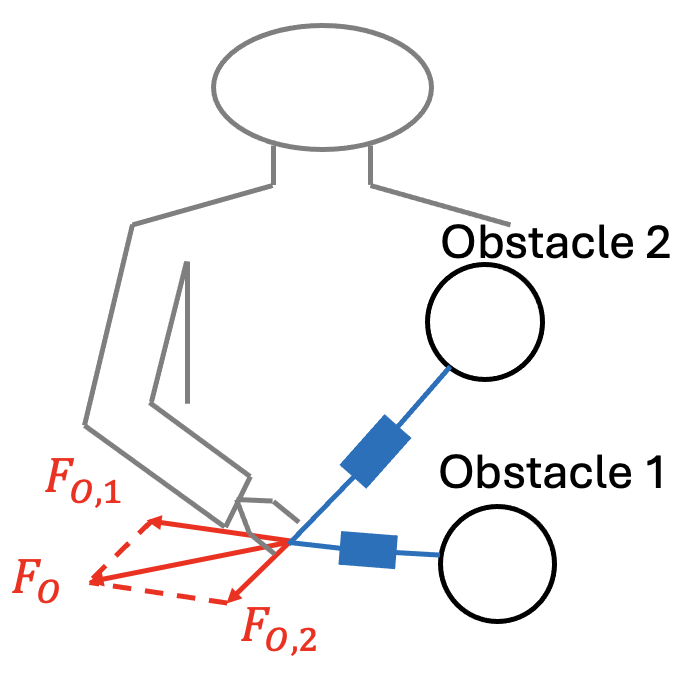}
		\caption{Multi-obstacles avoidance.}
		\label{obstacle_avoidance2}
	\end{subfigure}\hfill
	\caption{(a)-(f) Virtual mechanisms designed for satisfying different planning goals.}
	\label{virtual_mechanisms}
\end{figure*}

Our hypothesis is that virtual model control offers a practical solution to the challenge of safe motion planning and control, by negotiating between various task specifications in complex environments.  
Specifically, this paper proposes six distinctive ways to design virtual models for the purpose of reaching in uncertain cluttered environment, illustrated in Fig. \ref{virtual_mechanisms}.
\emph{Simple reaching} is the most intuitive motion planning approach.
It shows similarities with PD control and impedance control in end effector space \cite{7110619,2023SPIE12645E}.
\emph{Geometric}, \emph{moving-reference}, and \emph{force-driven
reaching} extend motion planning with additional features, leading to 
a higher degree of accuracy and design flexibility. 
\emph{Obstacle} and \emph{multi-obstacles avoidance} consider how to reach the target location while avoiding known obstacles.

We highlight two key features of virtual model control:
(i) \emph{predictability} of (controlled) behavior and 
(ii) its \emph{intuitive design}.
Virtual model control makes the action of the controller explicit. 
The resulting control algorithms are predictable because they correspond to a mechanical interaction driven by physics. By inspecting the virtual mechanisms and the environment where the robot operates, we have an expectation of the intended motion, which makes the controller easier to understand and debug. Furthermore, given the planning requirements, the design of the controller is as simple as placing virtual components and tuning their parameters.  This allows one to tailor the design of the controller to the specific features of the task.
Through suitable tuning of the (virtual) mechanical parameters, the robot remains compliant.
The force exerted when a reaching cannot be achieved is controlled by the virtual stiffness of virtual mechanical springs. That is, virtual model control trades motion accuracy for controlled sensitivity to the interaction with the environment.  The controlled robot also enjoys the benefits of passivity \cite{Ortega2001}, i.e. stability and robustness, and controlled impedance \cite{Hogan1985a}. 

The paper is organized as follows. The virtual model control approach is discussed in Section \ref{methodology} through examples. The robot platform and implementation details are provided in Section \ref{implementation}. 
The experiments are discussed in Section \ref{experiments}: 
(i) unperturbed reaching, 
(ii) reaching with known obstacles, and 
(iii) reaching with unknown obstacles. Further discussions and conclusions are provided in Section \ref{conclusions}.

\section{Virtual Model Control} \label{methodology}

\subsection{Virtual model, virtual forces, and motor torques}

A virtual model controller is a collection of springs and dampers organized in a mechanical structure and connected to the robot. 
In this section, we show examples of virtual model controllers
for reach motion and obstacle avoidance.
Each controller is a virtual mechanism generating driving virtual forces, which are then translated and realized by robot actuators using torque control.
The translation can be obtained using the robot geometry.

For fully actuated robots, using $\pmb{q}$ to denote the joint coordinates of the robot, the force $\pmb{f}_i$ generated by a virtual component at a point $\pmb{h}_i(\pmb{q})$ of the robot is mapped into corresponding joint torques $\pmb{\tau}$ using the Jacobian matrix $J_i(\pmb{q}) = \frac{\partial \pmb{h}_i(\pmb{q})}{\partial \pmb{q}}$.
Following \cite{Spong2005}, the torque command to generate all the forces of the virtual model controller is given by 
\begin{equation}
\label{eq:translation}
    \pmb{\tau} = J_1(\pmb{q})^T \pmb{f}_1 + J_2(\pmb{q})^T\pmb{f}_2 + J_3(\pmb{q})^T\pmb{f}_3 + \dots
\end{equation}
Virtual model control is inherently stable, since the controlled robot is just the
interconnection of the passive mechanics of the robot with the passive virtual mechanics.
Stability is guaranteed by passivity theory \cite{Willems1972,VanDerSchaft1999,Ortega2023pid}. In fact, virtual model control provides an intuitive, task-oriented method to do energy shaping and damping injection in robotics \cite{Ortega2001,Ortega1998}. The selection of spring and damper parameters and their geometric configuration also shape the impedance of the robot \cite{Hogan1985a,Colgate1988}.

\subsection{Simple, geometrical, and force-driven reaching}
In its simplest form, moving the end effector to a desired position $\pmb{x}_d$ is achieved by a spring with stiffness $k$ and a damper with damping $c$ attached from the end effector position $\pmb{x}_{EE}$ to the desired position. 
These components produce the virtual force 
\begin{align}
	\pmb{f}_{EE} &= k\pmb{x} + c\dot{\pmb{x}}, \ \pmb{x} = \pmb{x}_d-\pmb{x}_{EE}. 
	\label{simplereach}
\end{align}
when the end effector deviates from the desired position and velocity, as shown in Fig. \ref{goal_reaching}. This corresponds to the classical proportional derivative control in the end-effector space.
Orientation can be controlled in a similar way, through rotational virtual spring and dampers. 

In more complicated reaching scenarios, such as those involving geometric constraints, e.g., a circular trajectory, the virtual mechanisms can be designed as in Fig. \ref{goal_reaching_complicated}.
A spring and damper connect the end effector to a virtual L-shaped structure, which has mass and is free to rotate relative to the table, driven by additional spring and damper. 
The spring and damper parameters and the geometry and position of the virtual structure concur to determine the virtual force that drives the end effector towards a circular motion. Compared to Fig. \ref{goal_reaching}, 
Fig. \ref{goal_reaching_complicated} shows how virtual model control can be used for (compliant) constrained motion.
The resulting virtual force is given by
\begin{subequations}
\begin{align}
	\pmb{f}_{EE} = k \pmb{x} + c \dot{\pmb{x}}, \ x = g(\theta) - \pmb{x}_{EE} \\
	I \ddot{\theta} + c_{\theta} \dot{\theta} + k_{\theta} (\theta) = w(\theta,\dot{\theta}, \pmb{f}_{EE}),
\end{align}
\end{subequations}
where $\theta$ is the angle from table of the virtual L-shaped part with inertia $I$, $c_{\theta}$ and $k_{\theta}$ are rotational damping and rotational stiffness parameters, respectively. 
$g$ represents the effect of the geometry of the L-shaped part. $w(\theta, \dot{\theta}, f_{EE})$ is the torque on the L-shaped part induced by the robot.

To cover large distances, a `moving' target approaches
mitigates force exertion, as shown in  Fig. \ref{trajectory}.
That is,
\begin{subequations}
\label{eq:tracking1}
\begin{align}
	\pmb{f}_{EE} &= k\pmb{x} + c\dot{\pmb{x}}, \ \pmb{x} = \pmb{x}_d (t)-\pmb{x}_{EE} \label{traj_eq}  \\
	\pmb{x}_d(t) &= \pmb{x}_0 + s(t)(\pmb{x}_g-\pmb{x}_0). \label{position_update}
\end{align}
\end{subequations}
$\pmb{x}_0$ is the initial position of the end effector. The straight motion of the moving target $\pmb{x}_d(t)$ is regulated by the parameter $0 \leq s(t) \leq 1$, starting at zero and converging to one in finite time. The robot follows the moving target in a compliant way.

Finally, Fig. \ref{trajectory_complicated} offers a more effective control of the force that the end effector exerts at contact with the environment. 
A virtual cart is attached to the end effector through spring-damper components and is pulled with a predefined constant force along a straight trajectory. 
The virtual force on end effector depends on the cart's dynamics.
\begin{subequations}
\label{eq:tracking2}
\begin{align}
	\pmb{f}_{EE} &= k\pmb{x} + c \dot {\pmb{x}}, \ x = \pmb{x}_C - \pmb{x}_{EE} \label{traj_eq}  \\
	m \ddot{\pmb{x}}_C + b\dot{\pmb{x}}_C &= \pmb{f}_C + {w}(\pmb{x}_C, \dot{\pmb{x}}_C, \pmb{f}_{EE}),
\end{align}
\end{subequations}
where the second equation models the additional dynamics of the cart.
$\pmb{x}_C$ and $\dot{\pmb{x}}_C$ are cart position and velocity.
The cart has mass $m$, its motion is affected by viscous friction $b$, and it is pulled by the force $\pmb{f}_C$.
$\pmb{w}(\pmb{x}_C, \dot{\pmb{x}}_C, \pmb{f}_{EE})$ models the reaction force on the cart due to the robot.

In the presence of obstacles or obstructions that stop the motion of the end effector, 
the virtual mechanism \eqref{eq:tracking2} guarantees that the total force exerted by the robot is bounded by the cart's force $\pmb{f}_C$. When the end effector is obstructed, also the cart reaches a stop. 
This means that the `desired' motion is naturally paused and can be smoothly restored by removing the obstacle. 
These features show how the introduction of virtual model control is well adjusted to support interactions with a partially known environment.

\subsection{Obstacle avoidance and compliant interaction}
In its simplest form, obstacle avoidance is achieved by
a repulsive virtual spring and a virtual damper that connect robot and obstacle, as shown in Figs. \ref{obstacle_avoidance1} and \ref{obstacle_avoidance2}.
We consider nonlinear springs that become stiffer as their length reduces. Thus, in combination with other virtual mechanisms, this strategy allows to prioritize obstacle avoidance when the robot is near obstacles.

Focusing on the end effector for simplicity, the driving virtual force $\pmb{f}_{\mathrm{o}}$ due to several repulsive springs attached between $N$ obstacles and end effector is given by
\begin{subequations}
\label{obstacles}
\begin{align}
	\pmb{f}_{\mathrm{o},i} &= k_i(|\pmb{x}_i|)\pmb{x}_i + c_i(|\pmb{x}_i|)\dot{\pmb{x}_i}, \
	\pmb{x}_i = \pmb{x}_{EE}-\pmb{x}_{\mathrm{o},i} ,\label{eq:one_obstacle}\\
	\pmb{f}_{\mathrm{o}} &= \Sigma_{i=1}^{N} \pmb{f}_{\mathrm{o},i},
\end{align}
\end{subequations}
where  $\pmb{x}_{\mathrm{o},i}$ is the position of the $i$-th obstacle.
$k_i(|\pmb{x}_i|)$ and $c_i(|\dot{\pmb{x}}_i|_i)$ are the associated nonlinear stiffness and nonlinear damping, respectively.

The case of unknown or uncertain obstacles is more challenging. For the unknown interactions
of cluttered environments we need to take advantage of the controlled compliance of the robot.
Consider the scenario of an undetected obstacle (acrylic plate) shown in Fig. \ref{uncertain_cluttered_reaching}. The behavior of the robot is determined by the interaction of its mechanics, the virtual mechanism, and by the additional contact with the obstacle. In fact,  this  contact force adds up to the forces generated by the virtual mechanism, therefore we can design the virtual mechanisms to guarantee that the robot behavior will degrade in a physically predictable manner. In case of obstruction, for example, the static picture is that the end effector will stop at the balance between the force exerted by the virtual mechanism springs and the reaction force from the obstacle. 
The total force expressed by the robot is thus regulated by the stiffness of the virtual springs and by the mechanical properties of the obstacle. Indeed, a soft obstacle will have a different effect than a rigid material.

\section{Robot platform and control architecture} \label{implementation}

\begin{figure}[htbp]
	\centering
   \vspace{2mm}
	\includegraphics[width=0.8\columnwidth]{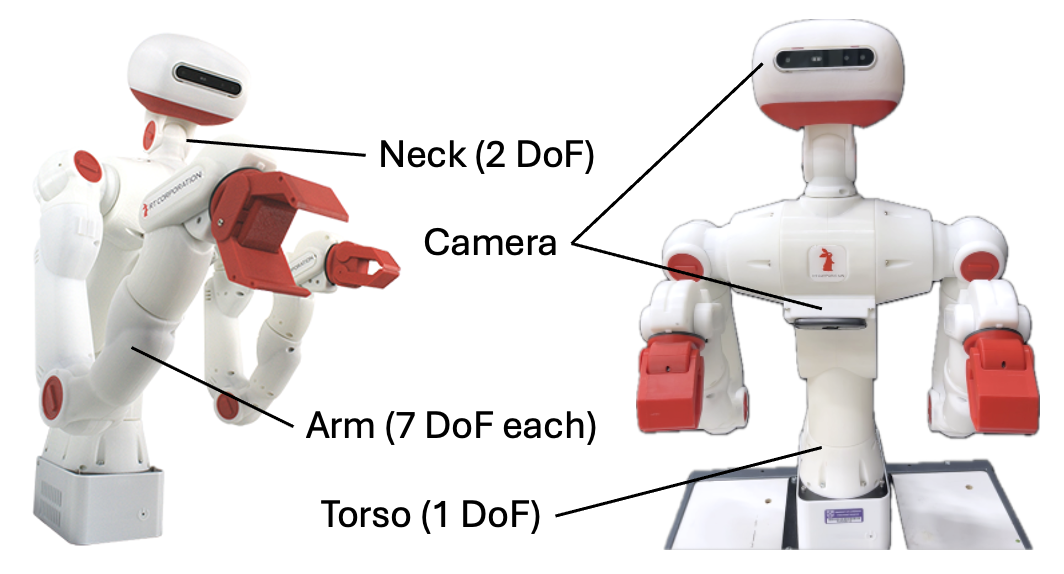}
	\caption{Experimental platform: Sciurus17 robot.}
	\label{Sciurus17}
\end{figure}

The experimental platform utilized for the study is the humanoid robot Sciurus17 shown in Fig. \ref{Sciurus17}, a lightweight low cost platform developed by RT Corp.
It has 17 degrees of freedom (DoF) distributed across its upper body, with seven DoF for each arm, two DoF for head and one DoF for torso.
The robot uses Dynamixel motors capable of current control. 
The conversion from torque to current is approximated as a linear equation using the performance graph of the Dynamixel motor.
The robot head is also equipped with an Intel RealSense D415 camera, which can be used to detect obstacles.

\begin{figure}[htbp]
	\centering
	\includegraphics[width=0.85\columnwidth]{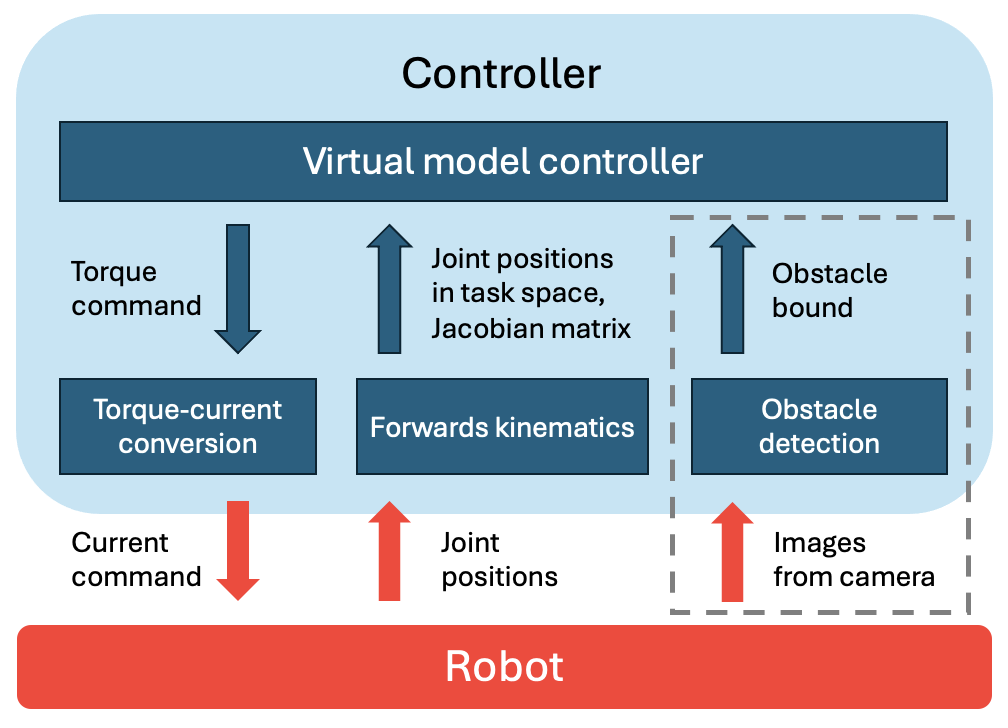}
	\caption{Control architecture.}
	\label{control_architecture} \vspace{-2mm}
\end{figure}

The controller runs on a standard computer.
The control architecture is depicted in Fig. \ref{control_architecture}.
The virtual forces generated by the virtual mechanisms are computed and translated into joint torques using \eqref{eq:translation}.
These are then converted to demanded currents using the approximated linear relationship.
The computation of the `state' of virtual mechanisms, such as the elongation of virtual springs and the velocities for the dampers, relies on position and velocity information of the robot joints, read from the Dynamixel motors. Inverse kinematics is not needed to control the robot. All forces and their translation into joint torques depend on forward kinematics and associated Jacobians.
The calculation involves only adding and multiplying matrices, as shown in \eqref{eq:translation}.
This is a significant departure from classical motion planning, whose 
execution often relies on inverse kinematics of the robot.

For obstacle detection, vision information such as image or point clouds can be captured from the camera. 
Once the obstacle is segmented, the nearest point on its bounding box or point cloud to the end effector can be easily identified. This point can then be used as the obstacle position in \eqref{eq:one_obstacle}.
However, for the experiments of this paper, vision-based obstacle detection is not used.
The position and radius of the obstacles are explicitly encoded.

\section{Experiments} \label{experiments}

\subsection{Reaching in obstacle-free environment}
\label{exp_reach}

\begin{figure}[htbp]
	\centering
		\includegraphics[width=0.8\columnwidth]{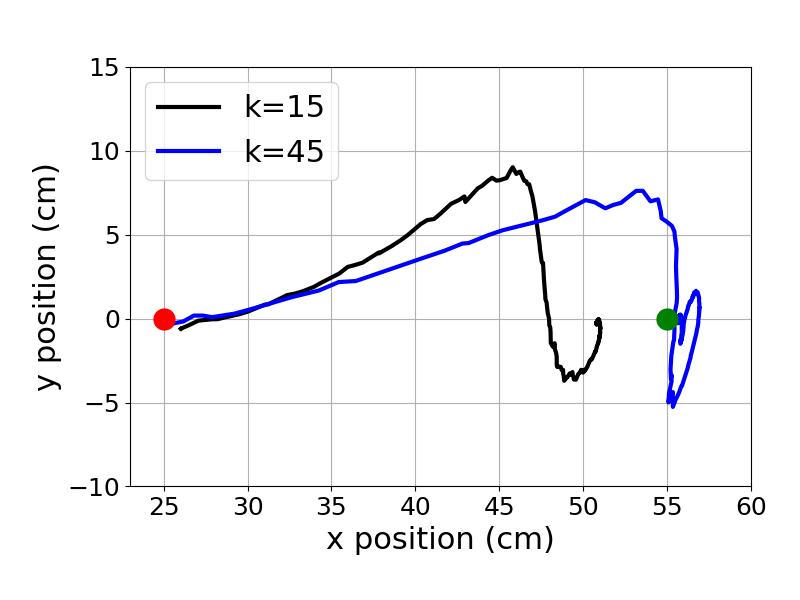} \hspace{-4mm}
		\hspace*{2mm}\includegraphics[width=0.8\columnwidth]{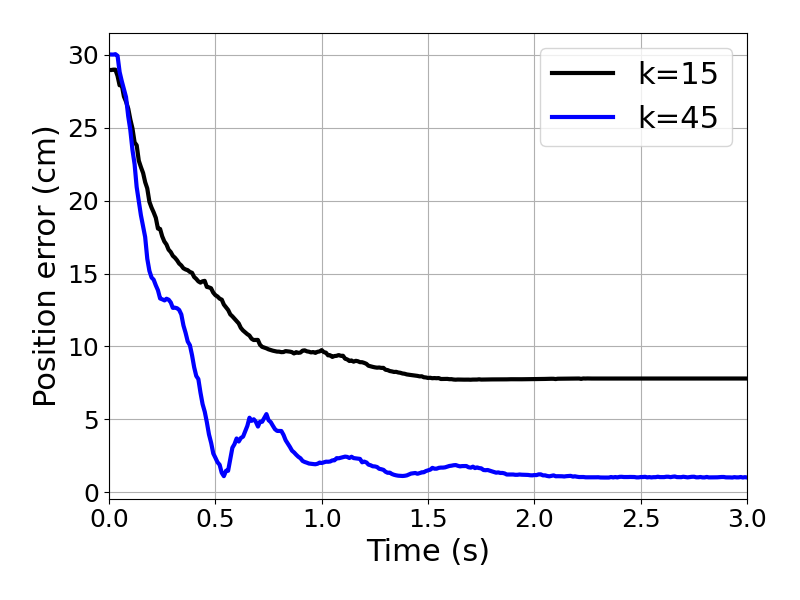} 
	\caption{\textbf{Simple reaching with different stiffness}. 
    Black line: $k=15$. \textcolor{blue}{Blue line}: $k=45$.
    \textcolor{red}{Red dot}: starting position. \textcolor{ForestGreen}{Green dot}: goal.
    Top: trajectory of the end effector on the x-y plane.
    Bottom: time-evolution of the reach error.}
	\label{exp1_reaching_onePoint} 
\end{figure}

Two reach motions are compared in Fig. \ref{exp1_reaching_onePoint}, both generated
using \emph{simple reaching} \eqref{simplereach}, for spring stiffness $k=15$ N/m and $k=45$ N/m while maintaining a constant damping coefficient of $1.5$ Ns/m.
It is not difficult to predict the behaviors for $k=15$ and $k=45$. The latter attains a smaller error and faster motion as shown in Fig. \ref{exp1_reaching_onePoint} (bottom).
In fact, for $k=15$ N/m, the force of the spring in the last section of the black trajectory is not enough to overcome friction.
During transients, the end effector tends to deviate along the y-axis before reaching the goal. This is the result of the interaction between the
virtual spring, the robot's arm inertia, and the inaccuracy of torque-current conversion for high-speed movements.

\begin{figure}[htbp]
	\centering
		\includegraphics[width=0.8\columnwidth]{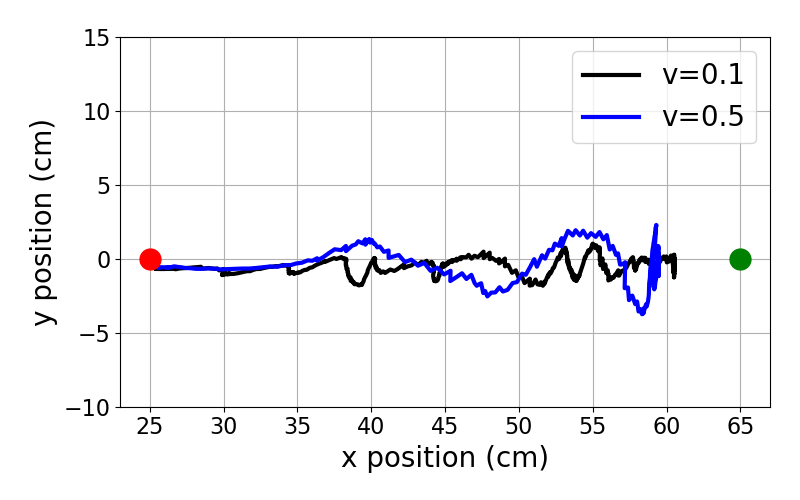} 
		\hspace*{1mm}\includegraphics[width=0.8\columnwidth]{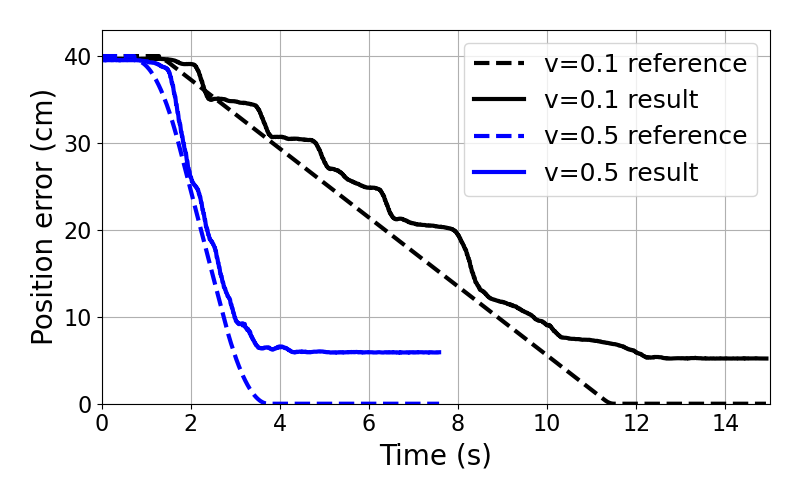} 
	\caption{\textbf{Moving-reference reaching with different speeds}. 
    Black line: $0.1$ m/s. \textcolor{blue}{Blue line}: $0.5$ m/s.
    \textcolor{red}{Red dot}: starting position. \textcolor{ForestGreen}{Green dot}: goal.
    Top: trajectory of the end effector on the x-y plane.
    Bottom: time-evolution of reference and actual distances from the target.}
	\label{exp2_reaching_onePoint} 
\end{figure}

Fig. \ref{exp2_reaching_onePoint} shows the trajectories obtained by
\emph{moving-reference reaching} \eqref{eq:tracking1}. Trajectories are
generated using spring stiffness $k=45$ N/m and damping coefficient of $1.5$ Ns/m.
The moving reference has a trapezoidal-shaped speed profile.
Compared to Fig. \ref{exp1_reaching_onePoint}, moving-reference
reaching limits high forces and fast motions.
The deviation along the $y$-axis is reduced. In fact, at lower speeds, the impact of imprecise torque-current conversion and robot inertia is diminished.
As in the previous case, the $5$ cm error from the target is due to the low stiffness of the reach motion. The force applied at the end of the trajectory does not overcome friction.

\subsection{Reaching while avoiding known obstacles}
\label{exp_obstacle_avoidance}

\begin{figure}[htbp]
	\centering
	\includegraphics[width=0.85\columnwidth]{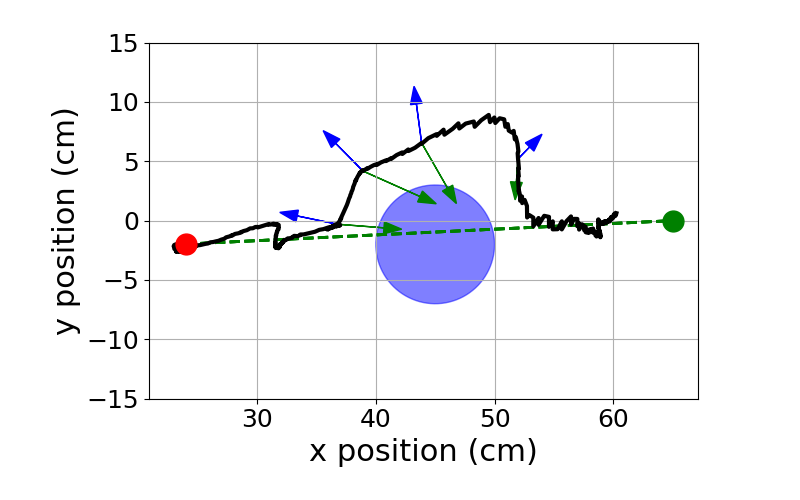}
	\includegraphics[width=0.83\columnwidth]{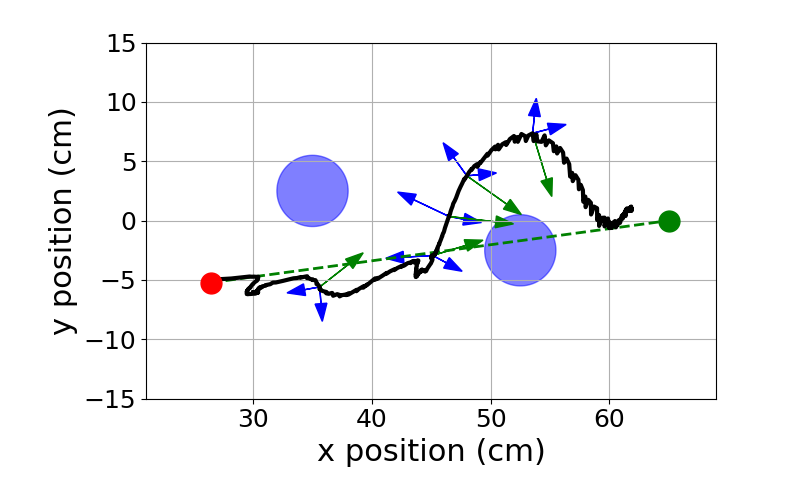} 
    \includegraphics[width=0.83\columnwidth]{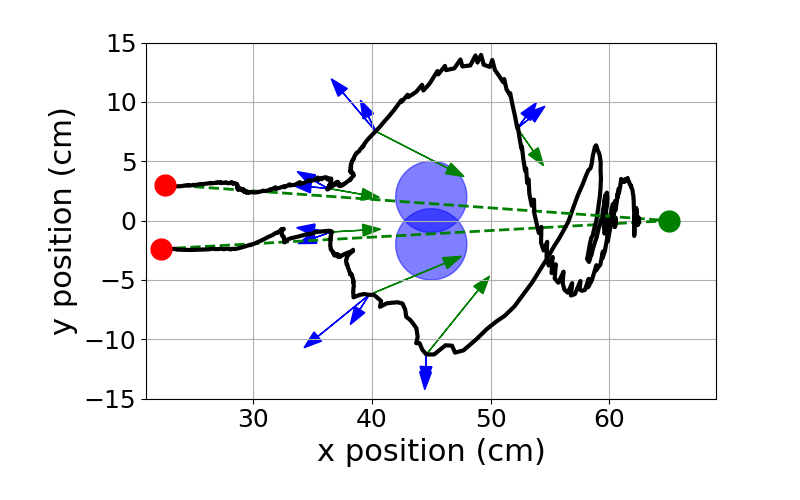}
	\caption{\textbf{Obstacle and multi-obstacle avoidance with known obstacles}. 
    Black line: robot trajectory. 
    \textcolor{blue}{Blue area}: cylindrical obstacles. 
    \textcolor{red}{Red dot}: starting position.
    \textcolor{ForestGreen}{Green dots}: ending positions. 
    \textcolor{ForestGreen}{Green line}: reference trajectory, 0.1 m/s.
    The arrows show the virtual mechanism forces. 
    \textcolor{blue}{Blue arrow}: force generated by obstacle avoidance virtual model. 
    \textcolor{ForestGreen}{Green arrow}: force generated by moving-reference reaching virtual model.} %Blue arrow: sum of both forces. }
	\label{cluttered_env_traj}
\end{figure}

A cylindrical obstacle with 10 cm diameter is placed %at (45, -2), 
as shown in Fig. \ref{cluttered_env_traj} (top). 
The black line shows the trajectory of the end effector in the x-y plane.
The robot is driven by moving-target reaching \eqref{eq:tracking1}, 
with speed 0.1 m/s, combined to \emph{obstacle avoidance} \eqref{obstacles}.
That is, for the latter, the obstacle is avoided through the action of a repulsive virtual nonlinear spring between the end effector and the obstacle. The stiffness of the virtual
spring is given by (in N/m)
\begin{align}
	0 \leq r < 5 \, \text{cm}, \quad k &= -46r + 250  \nonumber\\
	5 \, \text{cm} \leq r < 10 \, \text{cm}, \quad k &= -3.6r + 38  \nonumber\\
	10 \, \text{cm} \leq r < 20 \, \text{cm}, \quad k &= -0.2r + 4  \nonumber\\
        r \geq 20 \, \text{cm}, \quad k &= 0 \ \nonumber
\end{align}
where $r$ is the distance from end effector to the closest point on the obstacle.

The virtual model control offers a simple interpretation of the behavior of the robot. 
The end effector initially follows compliantly the straight-line reference. It is
then pushed away as it gets closer to the obstacle, the motion being determined by
the balance of forces between the repulsive spring from the obstacle and the attractive spring from the reference. Finally, after getting rid of the obstacle,
the end effector accelerates towards the goal. 

Two cylindrical obstacles with 6 cm diameter are then placed as shown in Fig. \ref{cluttered_env_traj} (middle). In this case, the obstacles are avoided by placing a repulsive virtual spring between end effector and each obstacle. 
It is not difficult to predict what the robot will do, as its motion
is now determined by the combined action of two repulsive springs.
In contrast to path planning algorithms,
the presence of two obstacles does not cause a significant increase in computation, 
as no pre-planning or re-planning are required.

As the last scenario, consider the obstacles of Fig. \ref{cluttered_env_traj} (bottom).
These partially overlap, forming a nonconvex shape. 
The resulting trajectories are not surprising.
The trajectories manage to avoid the obstacles after fluctuating for a while near the local minima of the elastic potential field formed by the virtual components. Similar to path planning with potential fields, this scenario shows how the combined effect of several springs may lead to issues. 
%depending on the initial condition. 
Indeed, this scenario calls for an improved design of the virtual model controller.

\subsection{Reaching while interacting with unknown obstacles}
\label{exp_obstacle_unknown}

The experiments of the previous sections show how different
virtual components can be combined to achieve complex goals. 
Both sections take advantage of available information to drive 
the robot behavior. In contrast, in this section we look
into uncertain interactions, given by undetected or inaccurately
positioned obstacles. 

\begin{figure}[htbp]
	\centering
		\includegraphics[width=0.7\columnwidth]{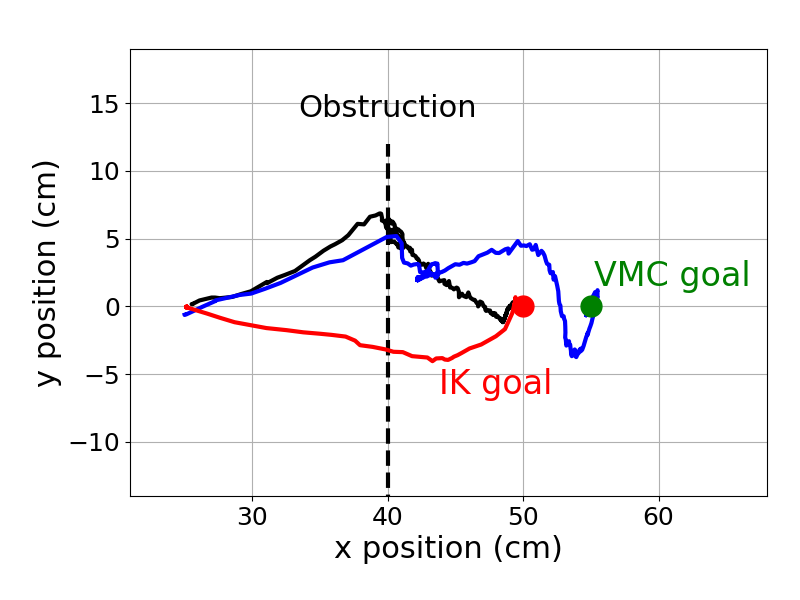} 
		\hspace*{2mm}\includegraphics[width=0.7\columnwidth]{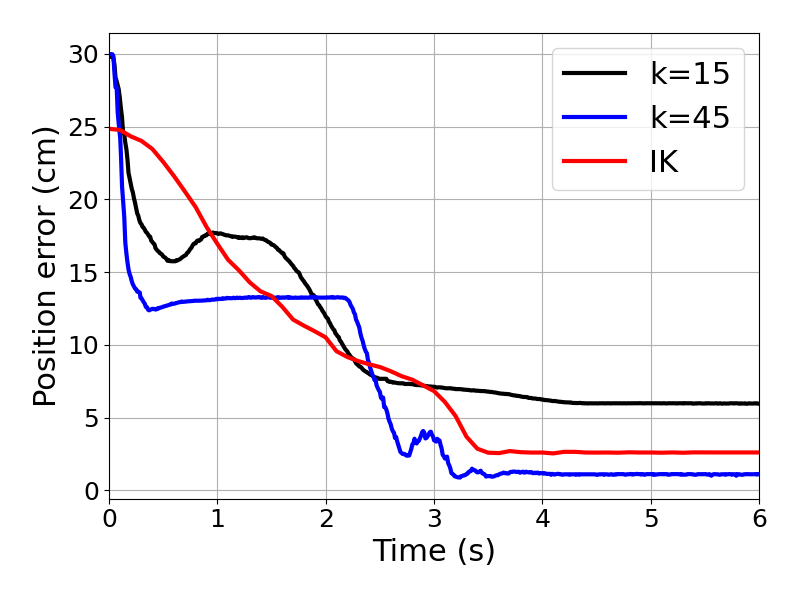}
 	\caption{\textbf{Simple reaching under undetected human interaction}.
    The end effector is blocked by a person at $x = 40$ cm.
    Black line: \eqref{simplereach}, $k=15$. 
    \textcolor{blue}{Blue line}: \eqref{simplereach}, $k=45$. 
    \textcolor{red}{Red line}: inverse kinematics.
    \textcolor{ForestGreen}{Green dot}: goal for VMC.
    \textcolor{red}{Red dot}: goal for inverse kinematics (which is different from VMC due to its inability to reach further using torso).
    Top: trajectory of the end effector on the x-y plane.
    Bottom: time-evolution of the reach error.}
	\label{human_robot_interaction} \vspace{-4mm}
\end{figure}

\begin{figure}[htbp]
	\centering
	\includegraphics[width=0.55\columnwidth]{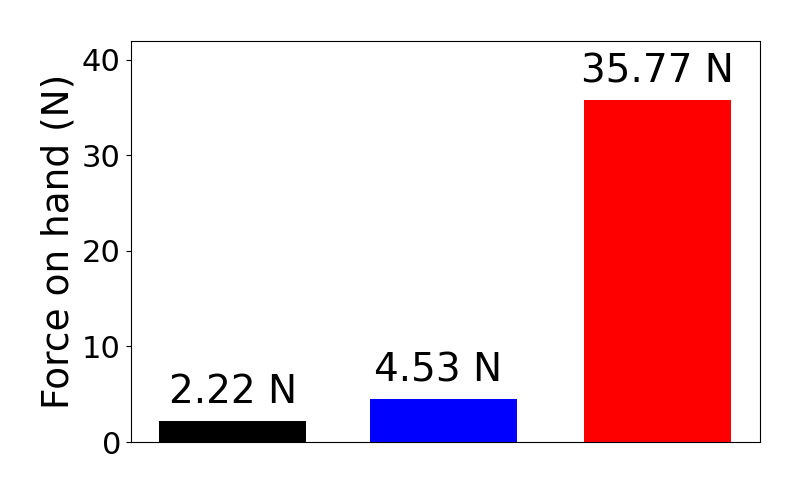}
	\caption{\textbf{Force exerted by end effector during obstruction}.
    Black: \eqref{simplereach}, $k=15$. 
    \textcolor{blue}{Blue}: \eqref{simplereach}, $k=45$. 
    \textcolor{red}{Red}: inverse kinematics.
    } %Force is measured when velocity is 0 for virtual model control cases
	\label{exp1_reaching_onePoint_force} \vspace{-5mm}
\end{figure}

As the first scenario, consider the reach task in Fig. \ref{exp1_reaching_onePoint},
based on the simple reaching virtual model controller \eqref{simplereach}.
This time, the end effector is blocked 
and held by a human hand in the middle of reaching, around $x = 40$ cm.
The robot has to face an unexpected (undetected) obstruction, as shown in Fig. \ref{human_robot_interaction} (top). The reaction of the robot is the one we expect
from the physical interpretation of the virtual model controller.
The end effector temporarily stops.
The force must be proportional to the stiffness of the spring and the distance to the target.
The motion of the robot resumes as soon as the end effector is released. 
The interaction with the human hand has no effect on the final error. 
For comparison, 
we have also implemented PID control in joint space based on inverse kinematics (IK). 
The differences with IK are straightforward. The virtual model controller applies
up to $4.53$ N force compared to $35.77$ N of the IK approach.
Contact forces are reported in Fig. \ref{exp1_reaching_onePoint_force}.

\begin{figure}[htbp]
    \centering
    \vspace{3mm}
        \includegraphics[width=.95\columnwidth]{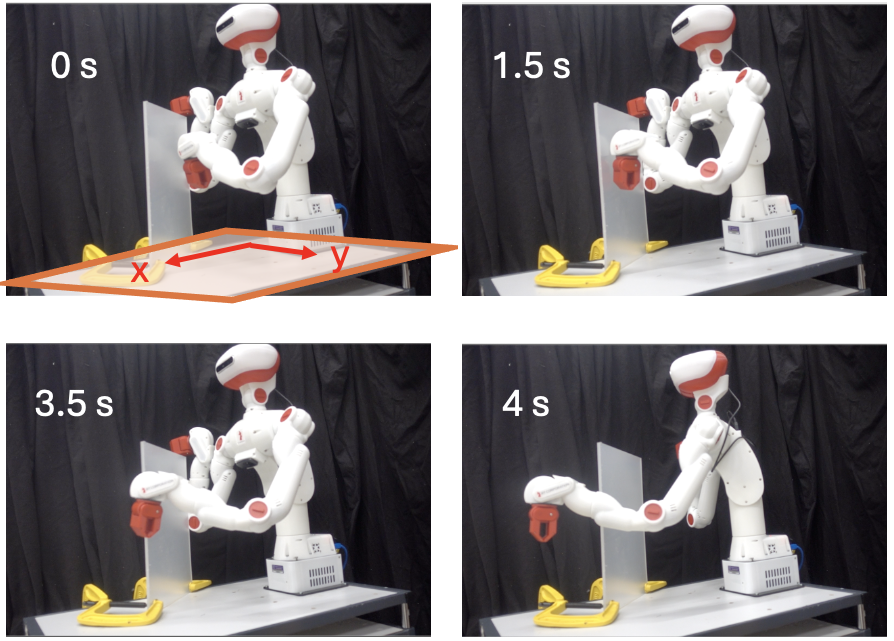}\vspace{5mm}
        \includegraphics[width=.75\columnwidth]{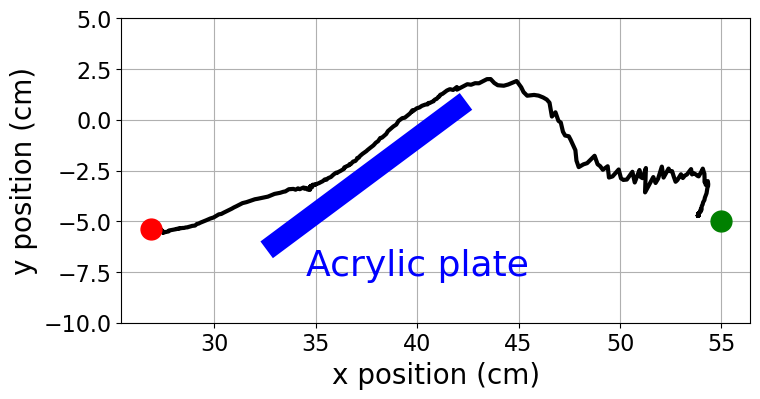} \vspace{2mm}
    \caption{\textbf{Moving-reference reaching under undetected obstacle}.
    Top snapshots: 
    $0$s - contact established, 
    $1.5$s - sliding, 
    $3.5$s - obstacle cleared,
    $4$s - goal reached. 
    Bottom figure: perturbed end effector trajectory.
    Black - trajectory, 
    \textcolor{red}{red dot} - starting position, 
    \textcolor{ForestGreen}{green dot} - goal, 
    \textcolor{blue}{blue} - acrylic plate.}
    \label{uncertain_cluttered_reaching}
\end{figure}

As the second scenario, consider the reaching task of Figure \ref{exp2_reaching_onePoint},
based on the moving-reference reaching virtual model controller \eqref{eq:tracking1}. 
The reference speed is $0.5$ m/s.  
This time, an acrylic plate is placed within the robot operative space and partially obstructs the motion, as shown in Fig. \ref{uncertain_cluttered_reaching}.
The presence and position of the plate are both unknown to the robot. 

Snapshots of the experiment and the resulting end effector trajectory are presented in Fig. \ref{uncertain_cluttered_reaching}.
After colliding with the obstacle, the end effector is held in place by the obstacle for around 0.5 seconds, then slides along it, and finally reaches the goal.
Virtual model control offers a simple explanation for this behavior.
Once in contact with the obstacle, the end effector is held in place until the virtual spring between the end effector and the moving reference extends enough to exert the required force to overcome friction and slide along the acrylic plate. The goal position is eventually reached once the obstacle is cleared. Notably, the `physical' perturbation of the acrylic plate shows similarities with the `virtual' perturbation of the obstacles of Fig. \ref{cluttered_env_traj}.

\section{Discussion and conclusions}\label{conclusions}

%\subsection{Related Works}
After its introduction by \cite{Pratt2001VirtualMC} for the control of walking bipedal robots, virtual model control remained underexplored in robot manipulation, even if several contributions in the literature show significant similarities.
In \cite{2003Brooks} virtual springs and dampers are used to control a robot into sequential patterns. Clear contact points can be found in the impedance control approach in end-effector space proposed by \cite{7506040,7110619}. In \cite{10.1007/978-3-642-34710-8_8} we see how virtual components are used to control the position of a physically simulated character's hand, posture, and walking cycles. Although these contributions recognize the value of using virtual components such as springs and dampers for control design, they do not venture into exploring the complexity and flexibility of virtual model control, with the introduction of additional dynamic and complex mechanical structures. This path is taken by \cite{LARBY20238548,Larby2022} 
to design a new controller based on a `virtual instrument' for robot laparoscopy. To the best of our knowledge, using virtual mechanisms for robot planning has not been explored in previous publications.

The examples of Section \ref{methodology} and the experiments of
Section \ref{experiments} support our hypothesis.
Virtual model control appears to have a significant advantage in terms of practical and intuitive design.
We design a virtual mechanisms for each task. Then, the robot can be controlled by activating the right collection of virtual mechanisms at run-time, as shown in the experiments. Section \ref{exp_obstacle_avoidance} shows that the controlled robot is predictable. The motion of the robot aligns with the expected interaction between the end effector, the environment, and the virtual mechanism.
The benefits of robot compliance are also evident. The interaction with the unknown obstacle of Section \ref{exp_obstacle_unknown} bears minor consequences on the reach motion. The reach motion is temporarily halted when a person grabs the robot end effector, but is restored right after.  In comparison with classical inverse kinematics we see how interaction forces are significantly reduced, enabling safer interactions.  

In virtual model control, 
virtual components, such as springs and dampers, respectively,
shape the potential energy of the controlled robot and the dissipation of the total energy. In this sense, for fully actuated robots, virtual model control provides a simple way to do energy-shaping and damping-injection \cite{Ortega1998,Ortega2001,Ortega2023pid}. The crucial difference is the design process. Representing the controller as a virtual mechanism allows to tailor the control action to the features of the task. In fact, consider the task of deriving an energy-shaping controller for reaching and avoiding obstacles. Finding the mathematical formula for the potential energy is a challenging task. In contrast, virtual model control guides the derivation of a desired potential energy, which is indirectly obtained as the combined energy of robot and virtual mechanism.

Interconnecting the robot with a compliant virtual mechanism is also
a way to shape the closed-loop impedance of the robot
\cite{Hogan1985a,Hogan1985b,Hogan1985c,7110619,Hogan2021}.  
In fact, a virtual mechanism with high stiffness and high damping coefficients
makes the robot stiff, thus insensitive to external forces. In contrast, low stiffness and low damping coefficients make the robot motion very compliant, thus sensitive to the interaction with the environment.
Although impedance control is a well-established and powerful methodology, finding the `right' impedance for a given task remains an elusive problem. 
In practice, impedance control too often reduces to the combination of dynamics inversion (to remove nonlinearities) and simple linear impedances. 
In contrast, virtual model control enables the design of complex nonlinear impedances, tailored to the task requirements. 

The purpose of this paper is to illustrate some fundamental features of virtual model control in dealing with integrated motion control and motion planning. Virtual model control provides an intuitive design approach for controlled robots that interact with the uncertain environment in a predictable way. Its design flexibility lies in the use of nonlinear mechanical elements organized in complex mechanical structures. Thus, virtual model control can take advantage of impedance compositionality, passivity, and additional compliant (virtual) dynamics to deliver robust and safe control for robots operating in unstructured and dynamic environments. That said, we are aware that this paper only scratches the surface of virtual model control design. 

Much remains to be understood. The choice of specific virtual mechanisms and the tuning of the virtual mechanical parameters were based on a trial-and-error approach. Optimal mechanisms and optimal parameter tuning are crucial research directions to address, with some initial results in \cite{Larby2022}. In a similar way, task prioritization in complex settings, the trade-off between precision and compliance, and the use of adaptive virtual mechanisms to optimize interactions at run time are key research directions to bring virtual model control to a mature design approach.

\addtolength{\textheight}{-12cm}   % This command serves to balance the column lengths

\bibliographystyle{IEEEtran}
\bibliography{bibliography}

\end{document}